\documentclass[letterpaper]{article} 
\usepackage{aaai23}  
\usepackage{times}  
\usepackage{helvet}  
\usepackage{courier}  
\usepackage[hyphens]{url}  
\usepackage{graphicx} 
\usepackage{natbib}  
\usepackage{caption} 
\usepackage{newfloat}
\usepackage{listings}

\usepackage{algorithm}
\usepackage{algorithmic}
\usepackage{times}
\usepackage{latexsym}
\usepackage[utf8]{inputenc}
\usepackage{microtype}
\usepackage{xspace}
\usepackage{caption}
\usepackage{subcaption}
\usepackage{amsmath}
\usepackage[nobiblatex]{xurl}
\usepackage{amssymb}
\usepackage{amsfonts}
\usepackage{graphicx}
\usepackage{tabularx}
\usepackage{arydshln}
\usepackage{mathtools,nccmath}
\usepackage{enumitem}
\usepackage{todonotes}
\usepackage{cleveref}
\usepackage{booktabs}

\DeclareCaptionStyle{ruled}{labelfont=normalfont,labelsep=colon,strut=off} 
\floatstyle{ruled}
\newfloat{listing}{tb}{lst}{}
\floatname{listing}{Listing}
\newcommand{\tf}[1]{\textbf{#1}}
\newcommand{\ti}[1]{\textit{#1}}
\newcommand{\ts}[1]{\textsc{#1}}
\newcommand{\tif}[1]{\textit{\textbf{#1}}}

\newcommand{\datasetname}{\ts{MultiSpider}\xspace}
\newcommand{\dam}{\ts{SAVe}\xspace}

\title{MultiSpider: Towards Benchmarking Multilingual Text-to-SQL Semantic Parsing}

\author {
    Longxu Dou\textsuperscript{\rm 1}, 
    Yan Gao\textsuperscript{\rm 2}, 
    Mingyang Pan\textsuperscript{\rm 1}, 
    Dingzirui Wang\textsuperscript{\rm 1}, \\
    Wanxiang Che\textsuperscript{\rm 1}, 
    Dechen Zhan\textsuperscript{\rm 1}, 
    Jian-Guang Lou\textsuperscript{\rm 2}
}
\affiliations {
    \textsuperscript{\rm 1} Harbin Institute of Technology\\
    \textsuperscript{\rm 2} Microsoft Research Asia\\
    \{lxdou, mypan, dzrwang, car\}@ir.hit.edu.cn, dechen@hit.edu.cn, \\
    \{Yan.Gao, jlou\}@microsoft.com
}

\begin{document}
\maketitle
\begin{abstract}

Text-to-SQL semantic parsing is an important NLP task, which greatly facilitates the interaction between users and the database and becomes the key component in many human-computer interaction systems.
Much recent progress in text-to-SQL has been driven by large-scale datasets, but most of them are centered on English.
In this work, we present \datasetname, the \ti{largest} multilingual text-to-SQL dataset which covers seven languages (English, German, French, Spanish, Japanese, Chinese, and Vietnamese).
Upon \datasetname, we further identify the lexical and structural challenges of text-to-SQL  (caused by specific language properties and dialect sayings) and their intensity across different languages.
Experimental results under three typical settings (zero-shot, monolingual and multilingual) reveal a $6.1$\% absolute drop in accuracy in non-English languages.
Qualitative and quantitative analyses are conducted to understand the reason for the performance drop of each language.
Besides the dataset, we also propose a simple schema augmentation framework \dam (\tif{S}chema-\tif{A}ugmentation-with-\tif{Ve}rification), which significantly boosts the overall performance by about $1.8$\% and closes the $29.5$\% performance gap across languages
\footnote{Code available at \url{https://github.com/microsoft/ContextualSP}}.

\end{abstract}

\section{Introduction}
Text-to-SQL semantic parsing is the task of mapping natural language sentences into executable SQL database queries, which serves as an important component in many natural language interface systems such as question answering and task-oriented dialogue. 
Despite the substantial number of systems~\cite{yin-neubig-2018-tranx, guo-etal-2019-towards, wang-etal-2020-rat, Scholak2021:PICARD} and benchmarks~\cite{yu-etal-2018-spider, yu-etal-2019-cosql,yu-etal-2019-sparc, guo-etal-2021-chase} for text-to-SQL, most of them are predominantly built in English, excluding this powerful tool's accessibility to non-English speakers.
The reason for this limitation lies in the serious lack of high-quality multilingual text-to-SQL datasets.

\begin{figure}[htp]
	\centering
	\includegraphics[width=0.45\textwidth]{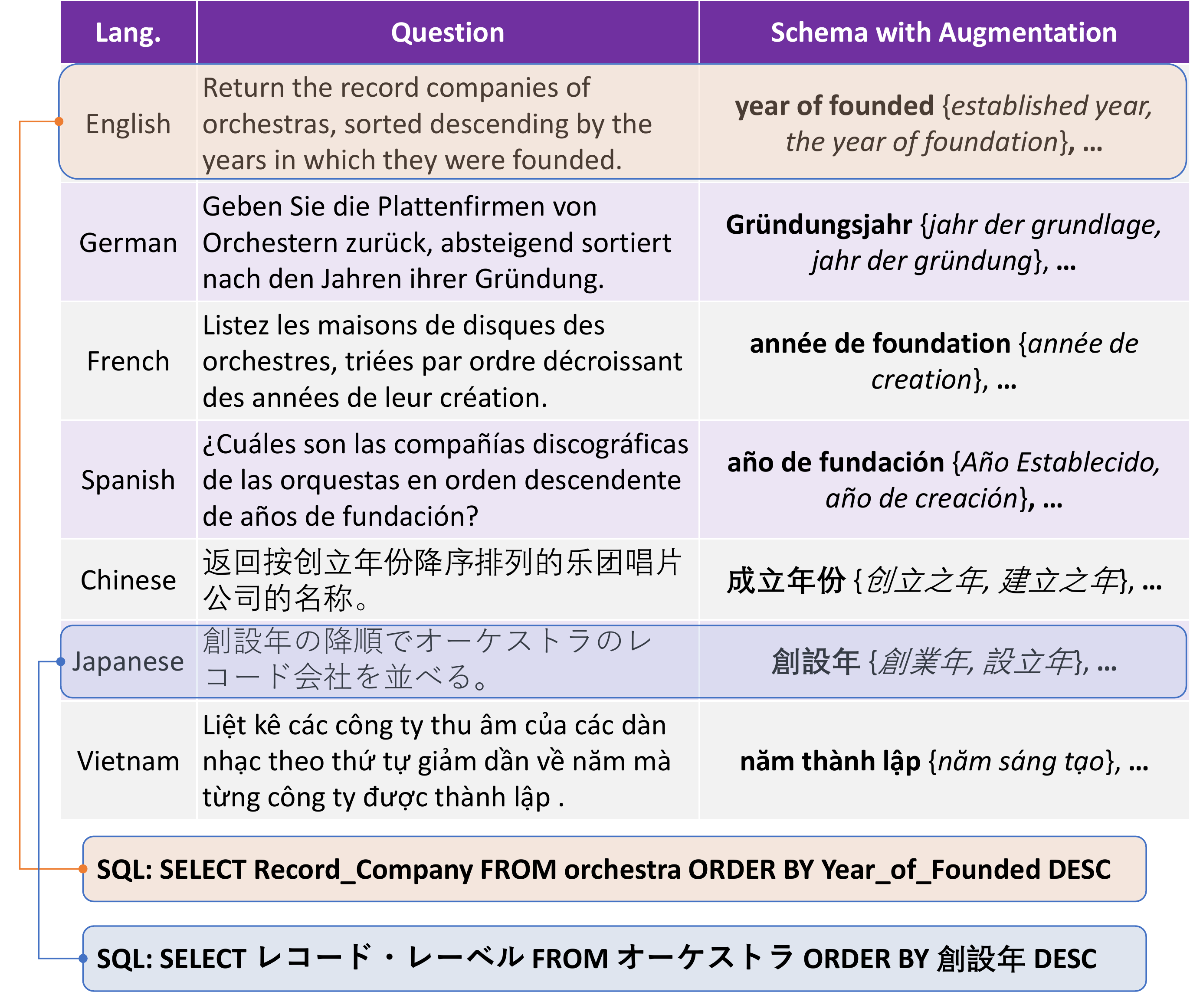}
	\caption{Examples of \datasetname and the augmented schema generated by \dam.} 
	\label{fig:mspider_example}
\end{figure}

Several works attempted to extend to new languages, but currently available multilingual text-to-SQL datasets only support three languages (English, Chinese, and Vietnamese) ~\cite{yu-etal-2018-spider,Min2019-CSPIDER,tuan-nguyen-etal-2020-vspider}, which hinders the study of multilingual text-to-SQL across a broad spectrum of language distances.
Besides the language coverage, the existing multilingual datasets also suffer from the following limitations:
\ti{(1)} \tif{low-quality}: unnatural or inaccurate translations; 
\ti{(2)} \tif{in-completed translation}: the database of CSpider~\cite{Min2019-CSPIDER} is not translated and still kept in English.
These limitations will inevitably lead to a limited multilingual system.
To advance multilingual text-to-SQL, in this paper, we present \datasetname, the \tif{largest} and \tif{high-quality} multilingual text-to-SQL dataset, which covers seven main-stream languages (Sec~\ref{sec:dataset_collection}).
Figure~\ref{fig:mspider_example} lists one example across seven languages including both question and schema.
To ensure the dataset quality, we first identify five typical translation mistakes during constructing a multilingual text-to-SQL dataset(Sec~\ref{sec:challenge_translation}), then we carefully organize the construction pipeline consisting of multi-round translation and validation (Sec~\ref{sec:pipeline}).
Most importantly, we take into account of the specific language properties to make the question more natural and realistic.  

Besides high-quality, \datasetname is quite challenging in multilingual text-to-SQL. Concretely, we explore the lexical and structural challenge~\cite{herzig-berant-2018-decoupling} of \datasetname (Sec~\ref{sec:data_analysis}): \ti{(1)} lexical challenge refers to mapping the entity mentions to schema alias (e.g., `record companies' to \ts{Record\_Company}); \ti{(2)} structural challenge refers to mapping the intentions to SQL operators (e.g., `sorted descending' to \ts{DESC}).
Experimental results and analysis demonstrate that \ti{(1)} the specific language properties like Hiragana and Katakana (Japanese) and morphologically rich language (German and French) make the lexical challenge more difficult by expanding the syntactic difference between schema and tokens; 
\ti{(2)} the dialect sayings require commonsense reasoning to address structural challenge (Figure~\ref{fig:dataset_analysis}).

To address the lexical challenges, we propose a simple data augmentation framework \dam from the view of schema,
which is more generic compared with the language-specific approaches \cite{Min2019-CSPIDER,tuan-nguyen-etal-2020-vspider} (e.g. PhoBERT for Vietnamese~\cite{phobert}).
Concretely, \dam consists of three steps (Sec~\ref{sec:save_pipeline}): 
\ti{(1)} conducting back-translation on contextualized schema using machine-translation;
\ti{(2)} extracting the the schema candidates;
\ti{(3)} measuring the semantic equivalency~\cite{pi-etal-2022-towards} with natural language inference model (NLI) to collect the suitable candidate. 
The quantitative and qualitative analysis prove that \ti{(1)} the augmented schema including synonyms and morphological variants; \ti{(2)} verification improve the accuracy of augmented data from $33.2$\% to $74.5$\% under human evaluation (Sec~\ref{sec:save_analysis_and_usage}).

To examine the challenge of \datasetname and verify the effectiveness of \dam, we conduct extensive experiments (Sec~\ref{sec:experiments}) under representative settings (zero-shot transfer, monolingual and multilingual). 
Experimental results reveal the absolute drop of accuracy in non-English languages is about $6.1$\% on average, indicating the difference in language causes the performance gap.
\dam significantly boosts the overall performance by about $1.8$\%, reducing the performance gap by $29.5$\% across languages.
We further study two research questions: \ti{what causes the performance drop in non-English languages?} (Sec~\ref{sec:case_studies}) and \ti{how schema augmentation \dam improves the model?} (Sec~\ref{sec:save_analysis}).

Our contributions can be summarized as follows:
\begin{itemize}
	\item  To our best knowledge, \datasetname is the \ti{largest}  multilingual text-to-SQL semantic parsing dataset with seven languages.
	\item We further identify lexical challenge and structure challenge of multilingual text-to-SQL brought by specific language properties. 
	\item We propose a simple-yet-effective data-augmentation method \dam from the perspective of schema.
	\item Experimental results reveal that \datasetname is indeed challenging and \dam significantly boosts the overall performance by about $1.8$\%. 
\end{itemize}

\section{The \datasetname Dataset}\label{sec:dataset_construction}
\subsection{Dataset Collection and Statistic}\label{sec:dataset_collection}
We build \datasetname based on Spider~\cite{yu-etal-2018-spider}, a large-scale cross-database text-to-SQL dataset in English. 
Only 9691 questions and 5263 SQL queries over 166 databases (train-set and dev-set) are publicly available. Thus we only translate those data.

Currently, there are two well-known extensions of Spider:
(1) CSpider~\cite{Min2019-CSPIDER} (Chinese, schema kept in English): we improve the existing translation and translate the schema as well.
(2) VSpider~\cite{tuan-nguyen-etal-2020-vspider} (Vietnamese): we re-partition the dataset to be the same as other languages for fair comparison.
The mentioned value (e.g., location and name) in question are kept in English, to be consistent with the database content.

\begin{figure}[tp]
	\centering
	\includegraphics[width=1\linewidth]{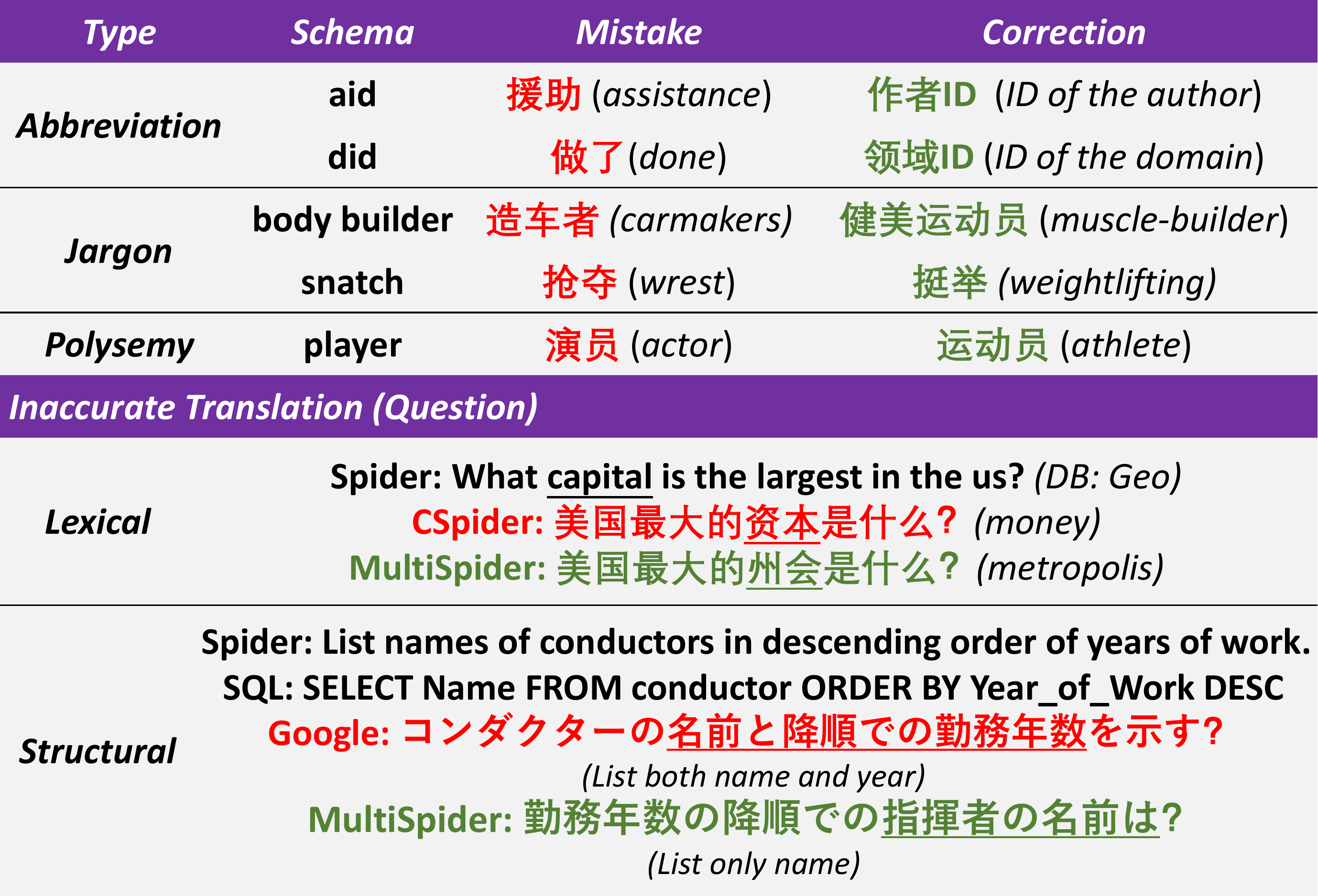}
	\caption{
	Typical mistakes (red, left/above) during the translation, due to the lack of context information and domain knowledge.
	The correct translation (green, right/below) and their explanations from WordNet.}
\label{fig:schema_translation_case}
\end{figure}

\begin{figure*}[htp]
	\centering
	\includegraphics[width=1\linewidth]{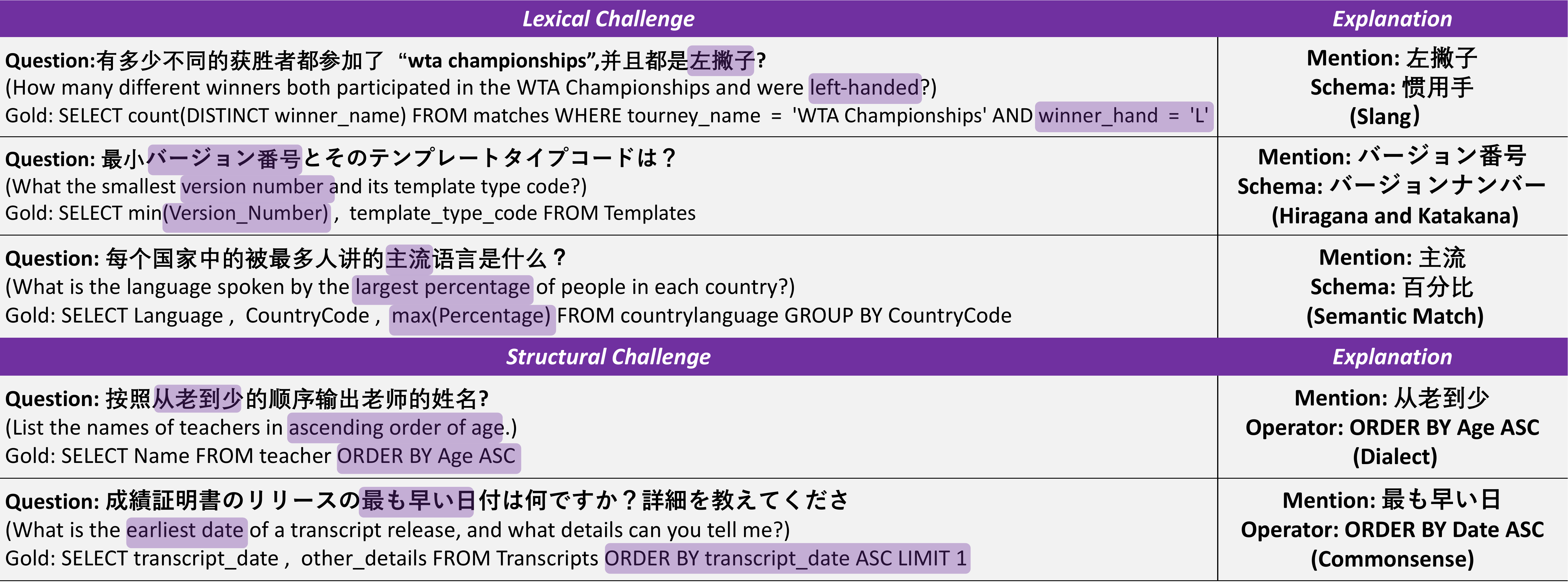}
	\caption{The lexical and structural challenge are further enhanced in \datasetname due to specific language properties.}
\label{fig:dataset_analysis}
\end{figure*}

\subsection{Challenge of Dataset Translation.} \label{sec:challenge_translation}
Based on our preliminary study, we summarize five typical mistakes during translating the text-to-SQL dataset including schema and question  (Figure~\ref{fig:schema_translation_case}).

\paragraph{Challenge of Schema Translation.} Both insufficient context and domain knowledge make the schema translation challenges, including abbreviation, domain-specific jargon, and polysemy. 
For example, \ts{aid} could be interpreted as `assistance' or `id\_of\_the\_author' (Figure~\ref{fig:schema_translation_case}).
We could disambiguate the meaning of schema headers by referring to its content value, neighbor headers, and involved question. 
Thus we can recognize \ts{aid} as the abbreviation of `id\_of\_the\_author' by examining its value `0001', neighbor `publisher', and the question `Return the aid of the best paper?'.

\paragraph{Challenge of Question Translation} 
We are facing two challenges here: 
\ti{(1)} lexical challenge refers to the entity polysemy , such as the `capital' in the case of Figure~\ref{fig:schema_translation_case}. 
It's not easy to deduce the actual meaning of `capital' (`money' or `metropolis') simply based on the context of the question, but we could disambiguate its meaning by schema translation of `capital' where domain knowledge is considered;
\ti{(2)} structural challenge points out that the complex logic or syntactic structure causes inaccurate translation.
We propose to refer to the corresponding SQL query to validate the logic.
For example, as shown in the last line of Figure~\ref{fig:schema_translation_case}, the machine translation might generate redundancy headers `year', while \datasetname avoids this.

\begin{figure}[tp]
	\centering
	\includegraphics[width=1\linewidth]{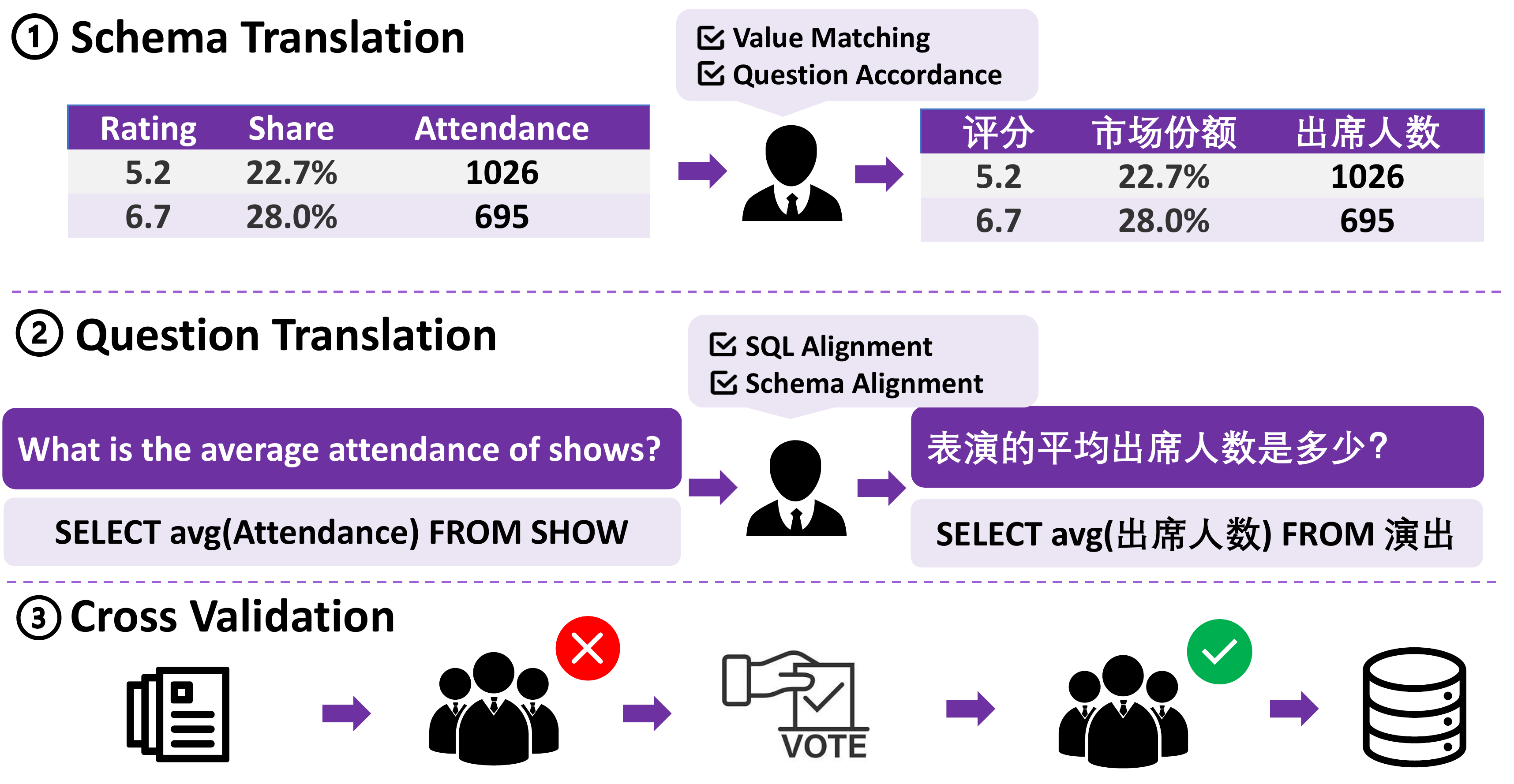}
	\caption{The translation pipeline of \datasetname, which consists of three steps: schema translation, question translation and cross-validation.
    }
\label{fig:translation_pipeline}
\end{figure}

\subsection{Translation Pipeline}\label{sec:pipeline}
\paragraph{Hiring Qualified Translators}
The translators are college students who majored in the target language\footnote{The payment of translators is listed in Ethical Statement.}.
There are three students for each language (15 students in total) who are proficient in English (e.g. IELTS $>=$ 7.0) and also meet the criteria: (1) language certificate of the target language, i.e, TEF/TCF for French; or (2) lived abroad for years. 

\paragraph{Translation and Validation}
To be effective, we first use Google NMT to translate the Spider, then let each translator post-edit the translation individually. 
According to the preliminary study about translation mistakes in Sec~\ref{sec:challenge_translation}, the translation pipeline is organized as three steps (Figure~\ref{fig:translation_pipeline}):
\ti{(1)} schema translation to let the translators leverage the content values of the corresponding schema, the neighbor headers, and the involved questions, to obtain sufficient context information of schema;
\ti{(2)} question translation by referring to the translated schema and translating the corresponding SQL simultaneously to valid the complex logic of the sentence; 
\ti{(3)} cross validation to merge the annotated data through voting the best translations among three annotators.

\begin{table}[tp]
	\centering
	\small
	\begin{tabular}{lccccccc}
		\toprule
		{} & \bf DE & \bf ES & \bf FR & \bf JA & \bf ZH   \\
		\midrule
		{\bf Question}  & {$4,607$}  & {$3,567$}  & {$4,723$} & {$4,092$} & {$989$}  \\
		{\bf Column} & {$1,248$}  & {$682$}  & {$1,382$} & {$1,601$} & {$1,469$}  \\
		{\bf Table} & {$362$} & {$225$} & {$327$} & {$470$} & {$670$} \\
		\bottomrule
	\end{tabular}
	\caption{The statistics of post-editing data for each language.	ZH starts from CSpider while the others are translated from Google NMT.
	} 
	\label{tab:statistic_of_modified_data}
\end{table}

\subsection{Dataset Analysis}\label{sec:data_analysis}
\paragraph{High Quality}
Although Google Translation reveals the excellent performance, the annotators further improve the data via post-editing about $37.1$\% questions and $27.3$\% schema as shown in Table~\ref{tab:statistic_of_modified_data}.
For each language, we spent more than 200 hours on translation and data review.
In this way, the inter-agreement of annotators reaches $92.7$\%\footnote{The inter-annotator agreement is calculated as the percentage of overlapping votes.}.

\paragraph{More Challenging}
Text-to-SQL usually tackles two kinds of challenges~\cite{herzig-berant-2018-decoupling}, i.e. the \ti{structural challenge} (mapping the mentions to schema) and the \ti{lexical challenge} (mapping the intentions to SQL operators). 
As illustrated in Figure~\ref{fig:dataset_analysis}, \datasetname poses both the lexical and structural challenge in the context of multilingual: \ti{(1)} the specific language properties like Hiragana and Katakana (Japanese) and morphologically rich language (German and French) would make the lexical challenge more difficult by expanding the syntactic difference between schema and tokens; 
\ti{(2)} the translation question which involved dialect sayings requires further commonsense reasoning to address the structural challenge (Figure~\ref{fig:dataset_analysis}). 
Thus, \datasetname is also challenging in multilingual besides high-quality.

\section{Schema Augmentation: \dam} \label{sec:data_augmentation}
Lexical challenge becomes more severe in multilingual settings due to different language properties(Figure~\ref{fig:dataset_analysis}). 
To address this problem, we propose \dam (\tif{S}chema-\tif{A}ugmentation-with-\tif{Ve}rification) to generate more schema variations, to improve the grounding ability of the parser~\footnote{We choose schema augmentation rather than question augmentation since it’s more efficient. The augment coverage of a single table modification includes all affiliated SQLs.}. 

Specifically, we first adopt machine-translation to generate the synonym candidates of schemas by multi-rounds back-translation. Then we use natural-language-inference model to select the semantic equivalency candidates, via measuring the entailment scores between schema and candidate (to ensure the data quality). Eventually, the augmented schemas would be used to expand the training data.

\begin{figure}[tp]
	\centering
	\includegraphics[width=0.74\linewidth]{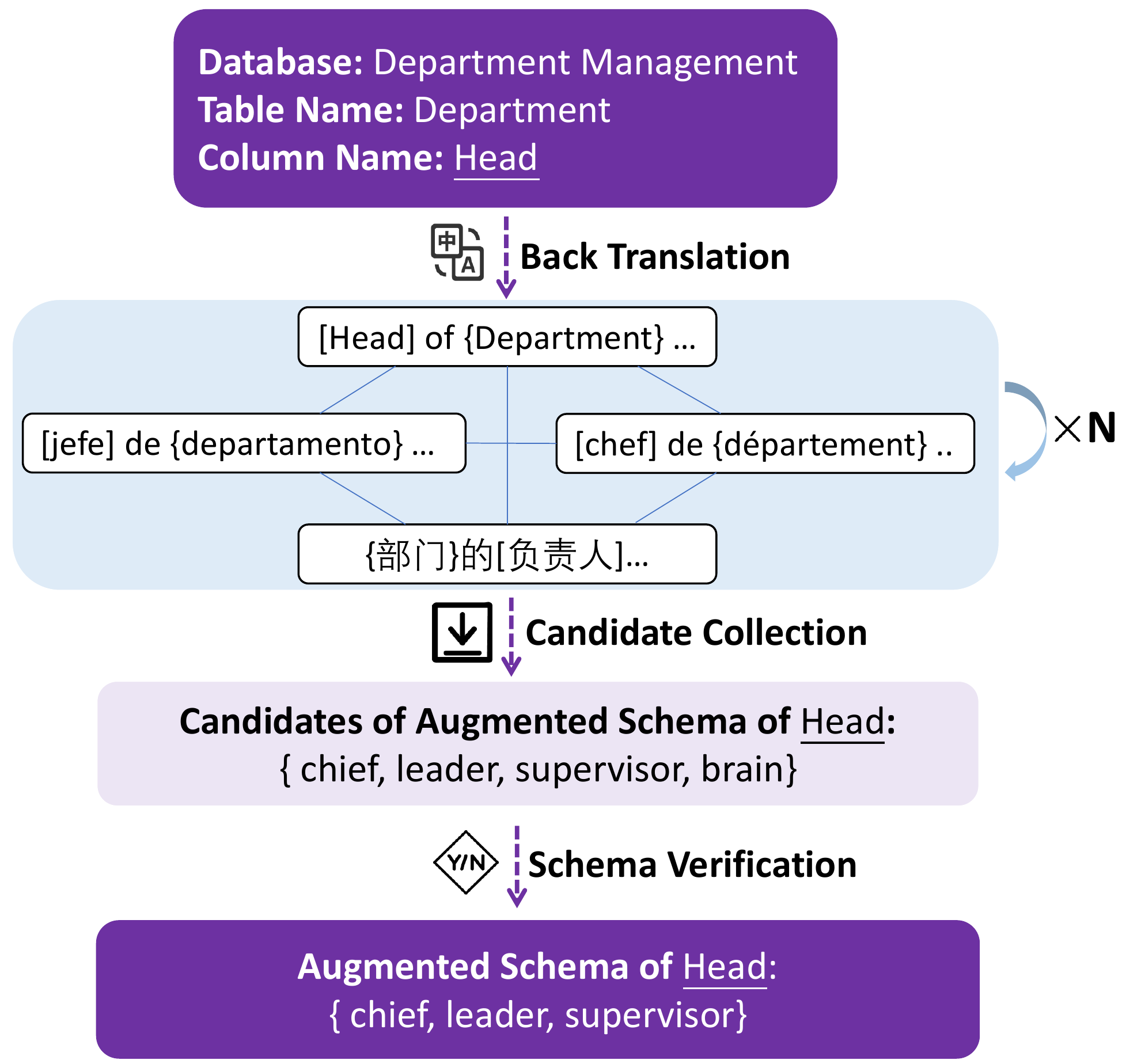}
\caption{The pipeline of Schema Augmentation.
	}
	\label{fig:augmentation_pipeline}
\end{figure}

\subsection{Augmentation Pipeline}\label{sec:save_pipeline}
\paragraph{Back Translation} generates the synonym candidates of schema (e.g. \ts{Chief} and \ts{Brain} are the candidates of \ts{Head}).
At first, to leverage the context of the schema for a better translation, we design a special template to insert the information of the database and the affiliated table like \ti{[COLUMN] of \{TABLE\} from (DATABASE\_NAME)}.
Then we translate this template from the target language into \ti{$K$} intermediate languages.
To further improve the candidate diversity, \ti{$N$} rounds of translation are conducted between intermediate language and target language alternatively.
Finally, we obtain \ti{$K*N$} synonym candidates (duplicate exists) in the target language\footnote{The back-translation would run $3$ turns among $11$ languages (seven languages of \datasetname plus Russian, Portuguese, Dutch, Swedish), i.e. \ti{$K = 11$, $N = 3$ }. The extra four languages are decided by their translations performance and the scale of their training corpus as reported in M2M100 paper.}. 

\paragraph{Schema Verification}
We propose to measure the semantic equivalence between the original schema and the candidate synonym to collect the suitable candidates inspired by~\citet{pi-etal-2022-towards}.
The main challenge in schema verification is to compute the similarity of contextualized schema (\ti{head of department vs. \ti{brain of department}}).
\citet{hill-etal-2015-simlex} shows that natural language inference (NLI) model achieves promising performance (70\% accuracy) than baselines (Word2Vec~\cite{word2vec} and Glove~\cite{pennington-etal-2014-glove}) in computing semantic equivalency. 
Thus, NLI model is a good choice to collect the synonym of schema via enumerating the candidates.
Concretely, we design a template to construct hypotheses and premise using schema and candidate as input.
The template is \ti{{TABLE} [COLUMN] (TYPE)} which contextualizes the schema with table context. 
Finally, we compute the entailment scores from both directions (\ti{premise to hypothesis} and \ti{hypothesis to premise}), as the judgment of semantic similarity. If they are both above the threshold ($0.68$ for Chinese and $0.65$ for others), we select this pair as augmented schema data.

\subsection{Quality of Augmented Schema}\label{sec:save_analysis_and_usage}
To examine the effectiveness of schema verification, we conduct the human-evaluation of the augmented schema.
Concretely, we sample 300 schemas from each language respectively.
The accuracy (i.e., the percentage of semantic equivalent items) is about $74.5$\% with verification and drops drastically to $33.2$\% without verification.

\subsection{Synthesising New Training Data}
The text-to-SQL data example consists of three parts: question, schema and SQL.
To expand the training corpus, for each data example, we randomly replace the schema items (e.g. COLUMN or TABLE) with the corresponding augmented schemas (e.g., replace \ts{Head} with \ts{Chief} in the above case) to compound the new training data examples. 
Consequently, we expand the training data by two to three times.

\begin{table*}[tp]
\begin{center}
\small
\begin{tabular}[b]{p{6cm}cccccccc}
    \toprule
    \bf {Model} & \bf EN & \bf DE & \bf ES & \bf FR & \bf JA & \bf ZH & \bf VI & \bf{AVG}(6 langs)\\

    \midrule
    \addlinespace
    \multicolumn{8}{l}{\textit{Monolingual Training (only use target language training data)}} \\
    \addlinespace
    \midrule
    mBART                  & $57.3$ & $39.7$ & $41.3$ & $37.5$ & $45.7$ & $55.0$ & $42.2$ & $43.6$\\
    mBART + \dam           & $58.3$ & $42.6$ & $42.6$ & $51.2$ & $46.9$ & $56.6$ & $43.1$ & $\tf{45.5 (+1.9\%)}$\\
     \hdashline 
    RAT-SQL + XLM-R        & $68.6$ & $62.5$ & $61.7$ & $64.1$ & $53.1$ & $63.4$ & $65.9$ & $61.8$\\
    RAT-SQL + XLM-R + \dam & $68.8$ & $63.9$ & $62.7$ & $65.7$ & $54.3$ & $66.2$ & $66.1$ & $\tf{63.2 (+1.4\%)}$\\    
            
    \midrule
    \addlinespace
    \multicolumn{8}{l}{\textit{Multilingual Training (use training data from multiple languages)}} \\
    \addlinespace
    \midrule
    mBART                  & $58.3$ & $42.7$ & $45.9$ & $42.9$ & $52.2$ & $57.8$ & $43.2$ & $47.5$\\
    mBART + \dam           & $59.7$ & $46.9$ & $47.1$ & $43.0$ & $54.3$ & $61.9$ & $45.6$ & $\tf{49.8 (+2.3\%)}$\\
     \hdashline 
    RAT-SQL + XLM-R        & $68.8$ & $64.8$ & $67.4$ & $65.3$ & $60.2$ & $66.1$ & $67.1$ & $65.2$\\
    RAT-SQL + XLM-R + \dam & $70.8$ & $66.7$ & $69.3$ & $67.5$ & $61.6$ & $67.3$ & $67.8$ & $\tf{66.7 (+1.5\%)}$\\

    \bottomrule
\end{tabular}
\caption{Exact-match Accuracy on \datasetname for 7 languages. 
Notice that the AVG is calculated across 6 non-English languages to be comparable to English results.
The performance boosts brought by \dam are bolded.
\label{tab:spider_results}}
\end{center}
\end{table*}

\section{Experiments}\label{sec:experiments}
\subsection{Experimental Setup}

\paragraph{Baseline Models}
We choose two types of representative models: (1) task-specific model RAT-SQL~\cite{wang-etal-2020-rat}, equipped with pretrained multilingual encoder mBERT~\cite{devlin-etal-2019-bert} and XLM-Roberta-Large~\cite{conneau-etal-2020-unsupervised-xlmr}; 
(2) pretrained multilingual encoder-decoder mBART~\cite{liu2020multilingualbart} which is inspired by the recent work of \citet{Scholak2021:PICARD} that reveals the excellent performance of pretrained encoder-decoder model. 

\paragraph{Evaluation Metric}
We report results using the same metrics as \cite{wang-etal-2020-rat}: exact match accuracy on all examples, as well as divided by difficulty levels determined by the official evaluation script~\cite{yu-etal-2018-spider}.

\paragraph{Training with Augmented Data}
During the training phase, we first adopt the augmented data to warm up the model three epochs to alleviate the noise in augmented data, then fine-tune the model with original high-quality training data.

\subsection{Experimental Results}\label{sec:results_and_analysis}
Follow the popular multilingual datasets MTOP~\cite{li-etal-2021-mtop} and MultiATIS++~\cite{xu-etal-2020-multiatis}, we conduct extensive experiments under three settings: zero-shot, monolingual and multilingual.
The results demonstrate that \ti{(1)} the absolute drop of accuracy in non-English languages is about $6.1$\% on average; \ti{(2)} \dam significantly improves the performance about $1.8$\% overall.

\begin{table}[htp]
\begin{center}
\small
\begin{tabular}[b]{lcccccc}
    \toprule
     \bf Model & \bf DE & \bf ES & \bf FR & \bf JA & \bf ZH & \bf VI\\
    \midrule
    \addlinespace
    \multicolumn{7}{l}{\textit{Directly Predict}} \\
    \addlinespace
    \midrule
    mBERT &  $50.9$ & $52.2$ & $50.7$ & $43.1$ & $49.6$ & $45.3$  \\
    XLM-R & $57.6$ & $60.8$ & $59.1$ & $48.3$ & $55.5$ & $56.5$ \\

    \midrule
    \addlinespace
    \multicolumn{7}{l}{\textit{Translate-then-Predict}} \\
    \addlinespace
    \midrule
    mBERT &  $49.6$ & $51.2$ & $47.6$ & $39.1$ & $46.7$ & $43.3$ \\
    XLM-R &  $58.8$ & $57.2$ & $58.7$ & $46.3$ & $55.3$ & $53.8$ \\
    
    \midrule
    \addlinespace
    \multicolumn{7}{l}{\textit{Translate-then-Train}} \\
    \addlinespace
    \midrule
    mBERT & $49.5$ & $51.2$ & $51.3$ & $38.2$ & $45.8$ & $49.3$  \\
    XLM-R & $\tf{60.2}$ & $\tf{61.9}$ & $\tf{61.7}$ & $\tf{51.3}$ & $\tf{57.6}$ & $\tf{63.9}$ \\

    \bottomrule
\end{tabular}
\caption{Exact-match Accuracy under zero-shot settings. 
}\label{tab:spider_zero-shot_results}
\end{center}
\end{table}

\subsubsection{Zero-shot Transfer.}
Zero-shot transfer is a realistic scenario where only the English training dataset is available.
We study three fine-grained zero-shot settings: 
\begin{itemize}
	\item \textbf{Directly Predict}: The parser is trained on English. During the inference, we directly predict with the question and schema in the target-language.
	
	\item \textbf{Translate-then-Predict}: The parser is trained on English. During the inference, we first translate the input question and schema from the target-language into English using Google NMT and then predict it. 
	
	\item \textbf{Translate-then-Train}: We first translate the original English dataset into the target language, then train the parser on this machine-translated training dataset. 
\end{itemize}

From Table~\ref{tab:spider_zero-shot_results}, we observed that 
\ti{(1)} the performance of zero-shot transfer largely depends on the choice of pre-trained encoder, where a better model enables better zero-shot transfer, i.e. XLM-R-Large beats mBERT by a large margin;
\ti{(2)} compared with translation-then-test, directly predict receives better performance about $1.6$\% since machine-translation might create mistakes, especially for schema translation;
\ti{(3)} with strong pretrained language model and machine-translation model, we could receive the promising results, which reveals that machine-translated data could be an economical proxy of human-translated data as \citet{sherborne-etal-2020-bootstrapping}. 

\begin{figure*}[htp]
	\centering
	\includegraphics[width=0.92\linewidth]{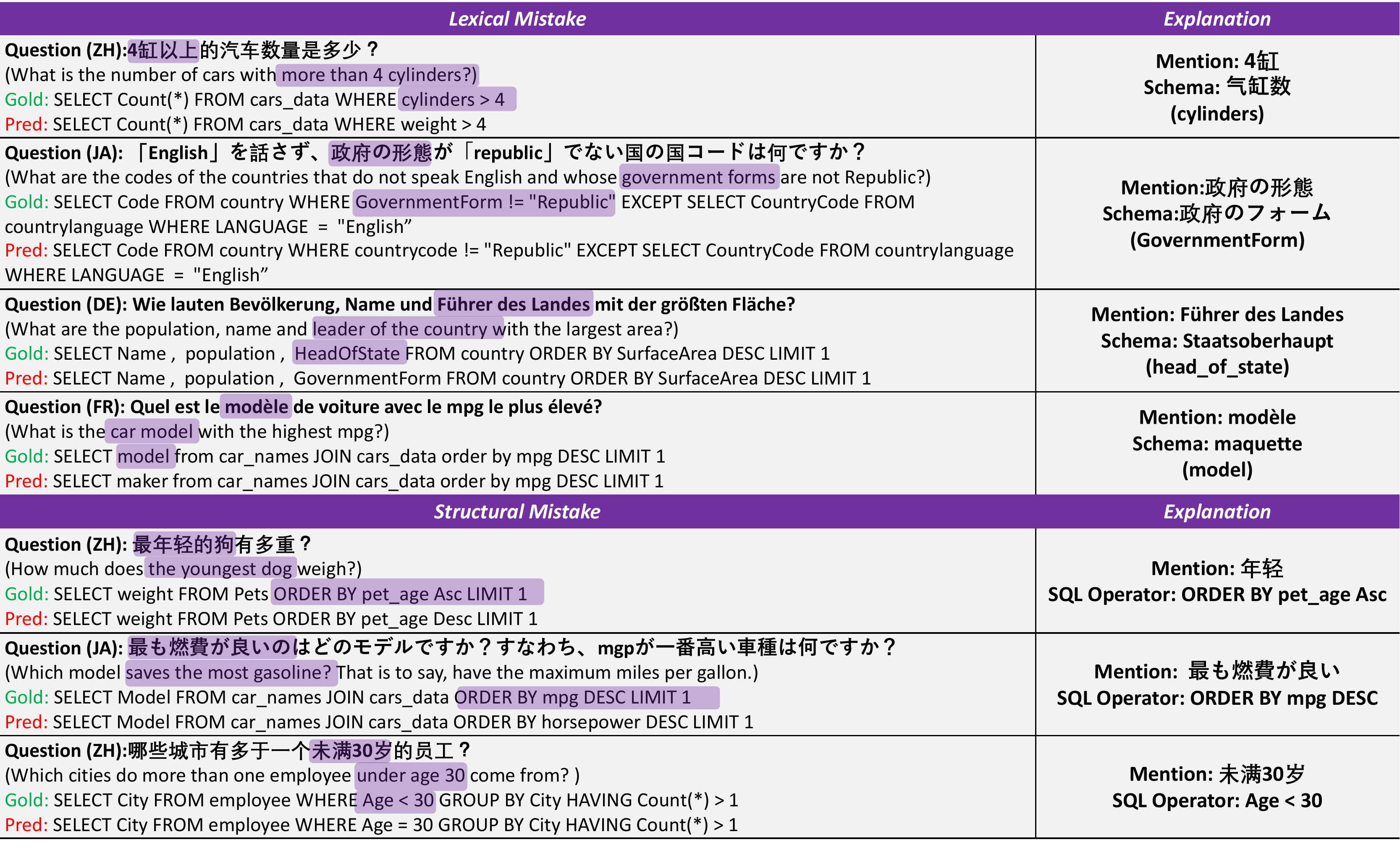}
	\caption{Case studies of non-English languages under two categories: lexical mistakes and structural mistakes.}
\label{fig:case_study}
\end{figure*}

\subsubsection{Monolingual Training.}
In this setting, the parser is trained on the human-translated training dataset in the target-language. 
From the results of the upper half of Table~\ref{tab:spider_results}, we observed that 
\ti{(1)} The performance of Japanese is significantly behind other languages. It's mainly caused by Hiragana and Katakana, which will be further analyzed in Sec~\ref{sec:case_studies};
\ti{(2)} BART exhibits strong performance in English and Chinese compared with the task-specific model, indicating the potential growth of pretrained seq2seq model in text-to-SQL;
\ti{(3)} \dam significantly improve the non-English languages (1.4\%-1.9\%) but raised less performance in English (0.2\%-1.0\%). 
We found that the most data pairs (schema and mention) in English are exactly/partly match~\cite{gan-etal-2021-spidersyn}, which is much easily than other languages so that it would benefit less from \dam.

\subsubsection{Multilingual Training.} 
In this setting, the parser is trained on the concatenation of training data from all languages.
From the results of the bottom half of Table~\ref{tab:spider_results}, we observed that 
(1) the multilingual training receives the best results overall. mBART and RAT-SQL receive a performance boost of about $3.9$\% from multilingual training in all languages; 
(2) English still benefits from multilingual training which is also proved by other multilingual datasets~\cite{xu-etal-2020-multiatis,li-etal-2021-mtop}; 
(3) Notably, \dam would improve the model further by 1.5\%, indicating the effectiveness of data augmentation.

\section{Discussion and Analysis}
\subsection{What causes the performance drop in non-English languages?} \label{sec:case_studies}
Sec~\ref{sec:results_and_analysis} demonstrates that the absolute drop of accuracy in non-English languages is about $6.1$\% on average. In this session, we attempt to conduct both qualitative analysis and quantitative analysis.

Concretely, we conduct case studies (Figure~\ref{fig:case_study}) for incorrect SQL prediction in non-English compared to the correct SQL prediction in English. All these SQL are predicted by RAT-SQL+XLM-R+\dam under multilingual settings, which is the SOTA model in experiments (Table~\ref{tab:spider_results}).
Furthermore, we divide these bad cases into two categories: lexical mistakes and structural mistakes. 

\begin{figure}[htp]
	\centering
	\includegraphics[width=0.83\linewidth]{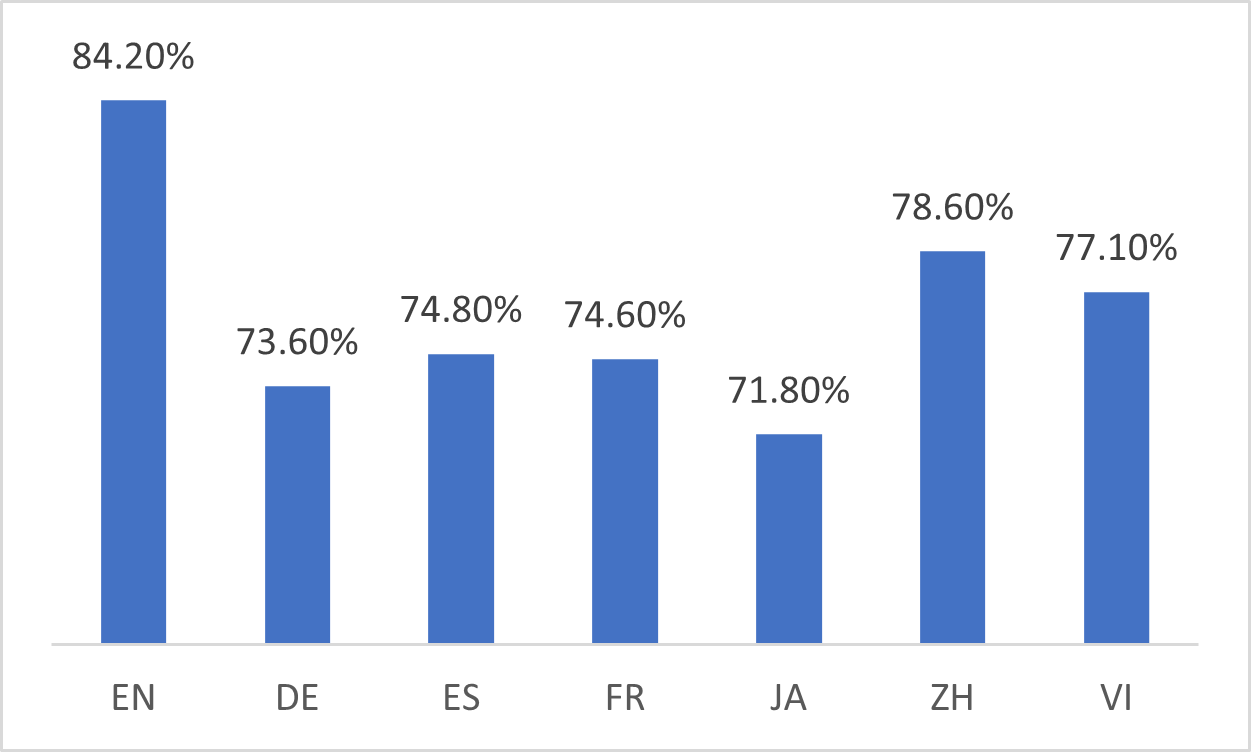}
	\caption{Fuzzy-match based schema-linking score, which measures the overlap between question and schema.} 
	\label{tab:schema-linking}
\end{figure}

\paragraph{Lexical Mistake} refers that the schema has not been grounded in SQL, which is usually caused by the syntactic difference between schema and tokens also known as schema-linking problem~\cite{wang-etal-2020-rat,lei-etal-2020-examining}. 
\ti{(1)} Qualitative analysis (Figure~\ref{fig:case_study}) reveals that the specific language properties like Slang (Chinese), Hiragana and Katakana (Japanese) and morphologically rich language (German and French) would expand the syntactic difference between schema and tokens then make the lexical challenge more difficult. In comparison, the original English Spider employ the similar surface form between entity mention and schema.
\ti{(2)} Quantitative analysis (Figure~\ref{tab:schema-linking}) computes the fuzzy-match-based score between question and schema, which is usually employed by popular model~\cite{wang-etal-2020-rat,guo-etal-2019-towards}, indicating that schema-linking becomes more challenging for non-English.

\paragraph{Structural Mistake} refers to the incorrect prediction of SQL operators.
The models are acquired to leverage the commonsense reasoning ability to match the SQL spans with intent mentions.
However compared with the English, \datasetname contains more dialect sayings in question annotation.
In the last case of Figure~\ref{fig:case_study}, 
it's quite a natural expression of `Age $<$ 30' in Chinese and non-trivial for model to deduce the actual meanings. 

In summary, both \ti{specific language properties} and the \ti{dialect sayings} lead to the performance drop in non-English languages, which also makes \datasetname more challenging in multilingual.

\begin{figure}[tp]
	\centering
	\includegraphics[width=0.95\linewidth]{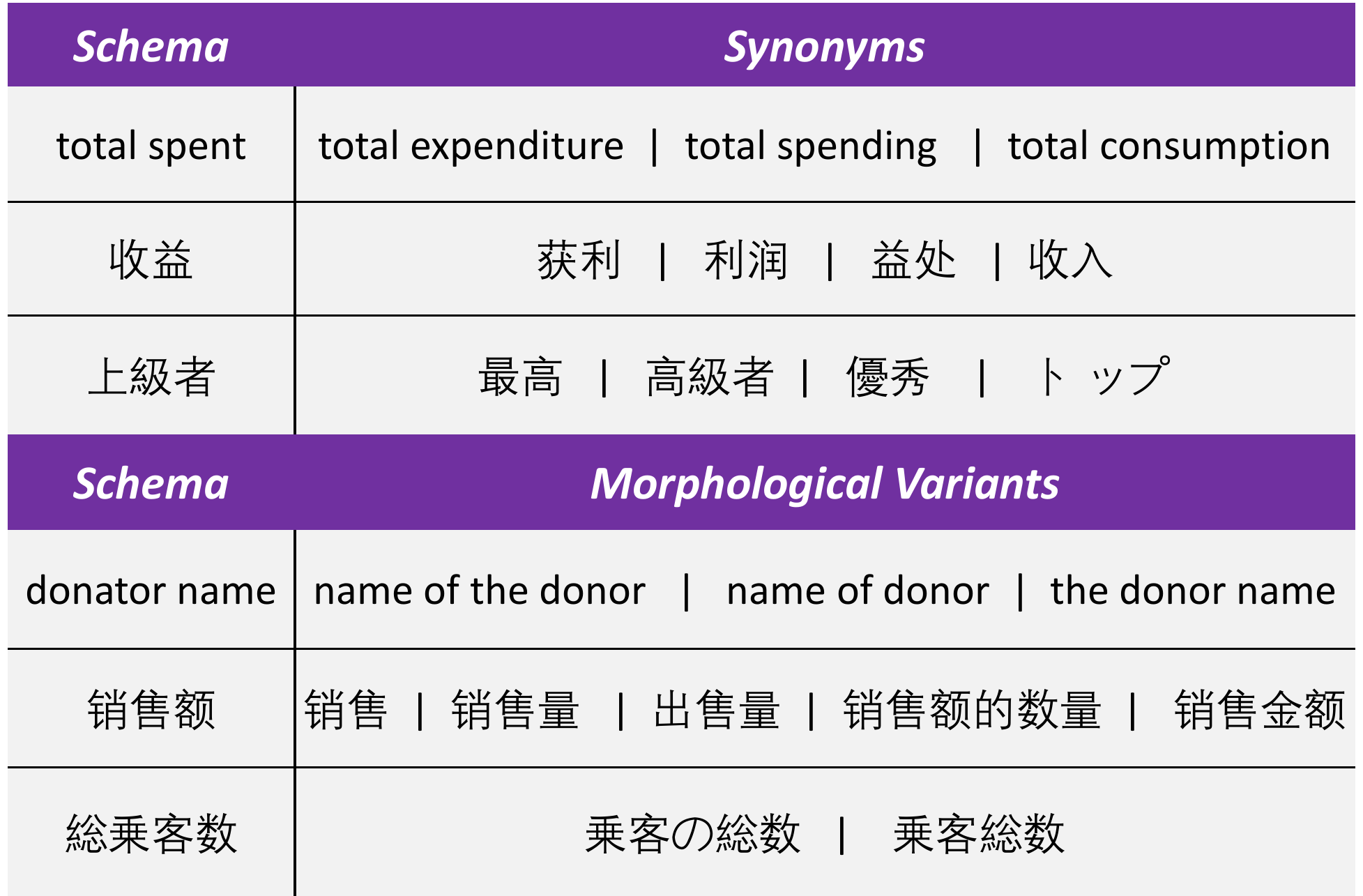}
	\caption{Two categories of augmented schema.
	}
\label{fig:save_analysis}
\end{figure}

\subsection{How schema augmentation \dam improves the model?}\label{sec:save_analysis}
Sec~\ref{sec:results_and_analysis} demonstrates that \dam significantly improves the performance by about $1.8$\% overall in all languages across three settings.
The performance gain of augmented data might comes from two aspects: 
\ti{(1)} addressing the lexical challenge by synthesizing more schema-token pairs;
\ti{(2)} improving the robustness of the text-to-SQL model through varies the schema input as studied by \cite{pi-etal-2022-towards}.

We conduct both qualitative and quantitative analysis on augmented schema to understand the reason of performance gain.
For qualitative analysis, after conducting cases studies on seven languages, we roughly classify the augmented schema items into two categories  (Figure~\ref{fig:save_analysis}):
\ti{synonyms} which is semantically identical with the original schema but with different lemmas (i.e. don`t have string overlap ); and
\ti{morphological variants} that changes the forms of schema syntactically.
For quantitative analysis, we sample 500 schemas from each language respectively, and we found that (1) for DE and ES, the most augmented schema (over 70\%) are morphological variants (2) for JA and ZH, it usually generate the synonyms.

\section{Related Work}
\subsection{Multilingual Text-to-SQL Datasets}
The recent development of text-to-SQL is largely driven by the large-scale annotation datasets.
These corpora cover a wide range of settings: single-table~\cite{zhongSeq2SQL2017-wikisql}, multi-table~\cite{yu-etal-2018-spider}, multi-turn~\cite{yu-etal-2019-cosql,yu-etal-2019-sparc}.
There are also a few non-English text-to-SQL datasets ~\cite{Min2019-CSPIDER,tuan-nguyen-etal-2020-vspider,guo-etal-2021-chase, Portuguese-spider}.

However, all these multilingual text-to-SQL datasets only support three languages.
The language coverage is limited compared with other multilingual datasets. 
For example, the multilingual task-oriented dialogue dataset MTOP~\cite{li-etal-2021-mtop} and MultiATIS++~\cite{xu-etal-2020-multiatis} support six languages and nine languages respectively.
Therefore, to advance the research on multilingual text-to-SQL, we propose \datasetname covering seven mainstream languages and quite challenging. 

\subsection{Multilingual Text-to-SQL Systems}
Driven by the large-scale English text-to-SQL dataset, many powerful task-specific model have been proposes for text-to-SQL, including effective input encoding~\cite{wang-etal-2020-rat}, intermediate representation of SQL~\cite{guo-etal-2019-towards} and grammar-based decoding for valid SQL~\cite{yin-neubig-2018-tranx}. 
Among a wide range of fancy models, RAT-SQL~\cite{wang-etal-2020-rat} is the most popular one which attracts a lot of attention from the research community and industry. Specifically, it adopts the relation-aware transformer to learn the joint representation of database and question, and achieves the promising results.

For non-English text-to-SQL, previous work \cite{Min2019-CSPIDER,tuan-nguyen-etal-2020-vspider} typically adopts language-specific tokenizer or pretrained language model like PhoBERT for Vietnamese~\cite{phobert}, to extend the English parser for multilingual scenario.
Therefore, we adopt the RAT-SQL with multilingual encoder like multilingual-BERT~\cite{devlin-etal-2019-bert} and XLM-R~\cite{conneau-etal-2020-unsupervised-xlmr} as our main baseline models. 

Besides the task-specific approaches, there is also another research trend that using the pretrained encoder-decoder models to track with the text-to-SQL.
It attempts to formula the text-to-SQL parsing tasks as seq2seq translation task.
Recently, researchers have developed lots of powerful parsers ~\cite{Scholak2021:PICARD, shin-etal-2021-constrained} built on the top of pretrained language models like BART~\cite{liu2020multilingualbart} and T5~\cite{2020t5}.
Thus, we attempt to choose mBART~\cite{liu2020multilingualbart}, a multilingual pretrained encoder-decoder model, as another baseline model.

\section{Conclusion and Future Work}
Most existing work on text-to-SQL are centered on English, excluding the powerful interaction technique's accessibility to non-English speakers.
In this paper, we present the \ti{largest} dataset \datasetname covering seven mainstream languages to promote the research on multilingual text-to-SQL. 
We ensure the dataset quality by hiring sufficient qualified translators and multi-rounds checking.
The results \datasetname is natural, accurate and also challenging in terms of text-to-SQL.
We further explore the lexical challenge and structural challenge in multilingual text-to-SQL and find that language-specific properties would make these two challenges more difficult. 
Therefore, we propose a simple and generic schema-augmentation method \dam to expand the size of training data.
Extensive experiments verify the effectiveness of \dam, which boosts the model performance by about $1.8$\%.
We propose a series of popular baseline methods and conduct extensive experiments on \datasetname to encourage future research for multilingual text-to-SQL systems.

Future work would include 
\ti{(1)} developing a multilingual text-to-SQL system and apply it in the real globalization scenario; 
\ti{(2)} leveraging better pretrained model and advancing architecture design to address the lexical challenge and structure challenge in multilingual settings;
\ti{(3)} expanding \dam to other table-related task~\cite{2019TabFactA} and further improve the schema verification accuracy.

\section*{Ethical Statement}\label{sec:ethical}
This work presents \datasetname, a free and open dataset for the research community to study the multilingual text-to-SQL problem.
Data in \datasetname are collected from Spider~\cite{yu-etal-2018-spider}, a free and open cross-database English text-to-SQL dataset.
We also collect data from the CSpider~\cite{Min2019-CSPIDER} and VSpider~\cite{tuan-nguyen-etal-2020-vspider}, which are also free and open text-to-SQL dataset.
To annotate the \datasetname, we recruit 15 Chinese college students (8 females and 7 males).
Each student is paid 2 yuan (\$0.3 USD) for translating the schema or questions.
This compensation is determined according to prior work on similar dataset construction~\cite{guo-etal-2021-chase}.
Since all question sequences are collected against open-access databases, there is no privacy issue.
The details of our data collection and characteristics are introduced in Section~\ref{sec:dataset_construction}.

\section*{Acknowledgement}
We thank all anonymous reviewers for their constructive comments. 
Wanxiang Che was supported via the grant 2020AAA0106501 and NSFC grants 62236004 and 61976072.
Dechen Zhan is the corresponding author.

\bibliographystyle{aaai23}
\bibliography{aaai23}

\end{document}